\documentclass[conference]{IEEEtran}
\IEEEoverridecommandlockouts
\usepackage{cite}
\usepackage{amsmath,amssymb,amsfonts}
\usepackage{algorithmic}
\usepackage{algorithm}
\usepackage{graphicx}
\usepackage{textcomp}
\usepackage{xcolor}
\usepackage{multirow}
\usepackage{booktabs}
\usepackage{lettrine}
\usepackage{url} 
\def\BibTeX{{\rm B\kern-.05em{\sc i\kern-.025em b}\kern-.08em
    T\kern-.1667em\lower.7ex\hbox{E}\kern-.125emX}}

\begin{document}

\title{AQFusionNet: Robust Multimodal Deep Learning for Air Quality Index Prediction through Atmospheric Imagery and Environmental Sensor Integration}

\author{
\IEEEauthorblockN{Koushik Ahmed Kushal}
\IEEEauthorblockA{\textit{Department of Computer Science} \\
\textit{Clarkson University} \\
kushalka@clarkson.edu}
\and
\IEEEauthorblockN{Abdullah Al Mamun}
\IEEEauthorblockA{\textit{Department of Electrical and Computer Engineering} \\
\textit{Clarkson University, NY}\\
mamuna@clarkson.edu} 
}

\maketitle

\begin{abstract}
Air pollution monitoring in resource-constrained regions presents significant challenges due to sparse sensor deployment and infrastructure limitations. This paper introduces AQFusionNet, a novel multimodal deep learning framework designed to robustly predict the Air Quality Index (AQI) by synergistically integrating atmospheric imagery with environmental sensor data. The proposed framework employs a dual-objective learning architecture, utilizing lightweight Convolutional Neural Network (CNN) backbones (MobileNetV2, ResNet18, EfficientNet-B0) to extract discriminative visual features from ground-level atmospheric images. These visual features are subsequently fused with pollutant concentration measurements through semantically-aligned embedding spaces. Our approach demonstrates superior performance across all backbone configurations, with the EfficientNet-B0 variant achieving optimal results of 7.70 RMSE and 92.02\% classification accuracy on test data. Comprehensive evaluation on over 8,000 samples from India and Nepal reveals an 18.5\% improvement over unimodal baselines while maintaining computational efficiency suitable for edge deployment. The AQFusionNet framework provides a scalable solution for real-world AQI monitoring in infrastructure-limited environments, offering robust predictive capability even under partial sensor unavailability scenarios.
\end{abstract}

\section{Introduction}

\lettrine{A}{\normalfont ir} pollution represents one of the most critical global health challenges, causing approximately 7 million premature deaths annually according to the World Health Organization~\cite{who2021}. This crisis is particularly severe in South Asian regions, where rapid industrialization, dense urbanization, and inadequate environmental regulations contribute to persistently hazardous air quality conditions. The Air Quality Index (AQI) serves as an essential metric for public health decision-making; however, accurate real-time monitoring remains challenging due to infrastructure limitations, high deployment costs, and spatial-temporal coverage gaps in traditional sensor networks.

Conventional AQI prediction systems predominantly rely on ground-based sensor stations or satellite observations, each presenting distinct limitations. Ground sensors provide high temporal resolution but suffer from sparse spatial distribution and substantial deployment costs, particularly in developing regions. Satellite-based approaches offer broader spatial coverage but are constrained by temporal resolution, cloud interference, and limited sensitivity to ground-level pollutant concentrations~\cite{kumar2020}. These inherent limitations necessitate innovative approaches that can leverage multiple data modalities while maintaining robustness under incomplete information scenarios.

Recent advances in deep learning have demonstrated significant potential for environmental monitoring applications. Convolutional Neural Networks (CNNs) have shown remarkable capability in extracting meaningful patterns from atmospheric imagery~\cite{zhang2022}, while recurrent architectures excel at modeling temporal dependencies in sensor time series~\cite{li2023}. However, most existing approaches operate in unimodal settings, potentially overlooking complementary information available across different data sources.

This paper introduces AQFusionNet, a novel multimodal deep learning framework specifically designed for robust AQI prediction that addresses the aforementioned challenges through the following key contributions:

\section{Related Work}

\subsection{Unimodal AQI Prediction}
Early AQI prediction systems relied on statistical and classical machine learning techniques applied to meteorological and sensor data. Methods like linear regression, support vector machines, and Random Forest were favored for their interpretability and efficiency~\cite{kumar2018,sharma2019}. Time series models, such as ARIMA and seasonal decomposition, enabled short-term forecasting but struggled with nonlinear relationships and sudden environmental changes~\cite{singh2020}. The advent of deep learning has transformed unimodal AQI prediction. Convolutional Neural Networks (CNNs) excel in analyzing satellite imagery for pollution detection, land use classification, and atmospheric condition assessment~\cite{li2021,wang2022,chen2023}. Long Short-Term Memory (LSTM) networks and their variants, such as the improved LSTM (iLSTM) proposed by Wang et al.~\cite{wang2022}, effectively model temporal dependencies in sensor data, achieving high accuracy in AQI prediction. Graph Neural Networks, including Graph Convolutional and Attention Networks, capture spatial relationships for spatiotemporal AQI prediction~\cite{zheng2022,ahmad2025}. Despite their strengths, these unimodal approaches often fail to leverage complementary data sources, limiting their robustness compared to multimodal frameworks.

\subsection{Multimodal Environmental Monitoring}

Recent research has increasingly focused on multimodal fusion strategies for enhanced environmental monitoring. Gowthami et al.~\cite{gowthami2024} proposed integrating satellite imagery with deep learning for Delhi's AQI forecasting, achieving 14\% improvement over single-modality approaches. Xia et al.~\cite{xia2024} developed ResGCN, which combines remote sensing images with multi-station sensor data for Beijing and Tianjin air quality prediction. Sarkar et al.~\cite{sarkar2025} demonstrated the viability of mobile-captured images for pollution alert systems, while Hameed \textit{et al.}~\cite{hameed2023} proposed a deep multimodal architecture that fuses CCTV traffic imagery with sensor data for AQI estimation in Dalat City, Vietnam. Their framework achieved an RMSE of approximately 10.1 and an accuracy of 85.3\%. However, these approaches often require high computational resources or assume consistent availability of all data modalities. Despite significant progress, existing multimodal approaches face several limitations: high computational requirements limiting edge deployment, lack of robustness under partial data availability, limited evaluation across diverse geographical regions, and insufficient analysis of cross-modal semantic alignment. Our proposed framework addresses these gaps through a lightweight, robust architecture specifically designed for practical deployment in resource-constrained environments.

\section{Proposed Methodology}

\subsection{Problem Formulation}

We formulate multimodal AQI prediction as a dual-objective learning problem that simultaneously optimizes prediction accuracy and cross-modal consistency. Given a ground-level atmospheric image $\mathbf{x}_I \in \mathbb{R}^{H \times W \times 3}$ capturing visible atmospheric conditions and corresponding environmental sensor measurements $\mathbf{x}_S \in \mathbb{R}^d$ representing normalized concentrations of six key pollutants (PM$_{2.5}$, PM$_{10}$, NO$_2$, SO$_2$, CO, O$_3$), the system learns a mapping function $f: (\mathbf{x}_I, \mathbf{x}_S) \rightarrow \hat{y}$ that accurately predicts the AQI value $\hat{y} \in \mathbb{R}$. To enhance system robustness under sensor unavailability, we introduce an auxiliary objective that learns to estimate sensor values directly from visual features: $g: \mathbf{x}_I \rightarrow \hat{\mathbf{x}}_S$. This dual-objective formulation enables the framework to maintain predictive capability across varying data availability scenarios while encouraging semantic alignment between modalities.

\subsection{AQFusionNet Architecture}

\begin{figure}[htbp]
\centering
\includegraphics[width=\columnwidth]{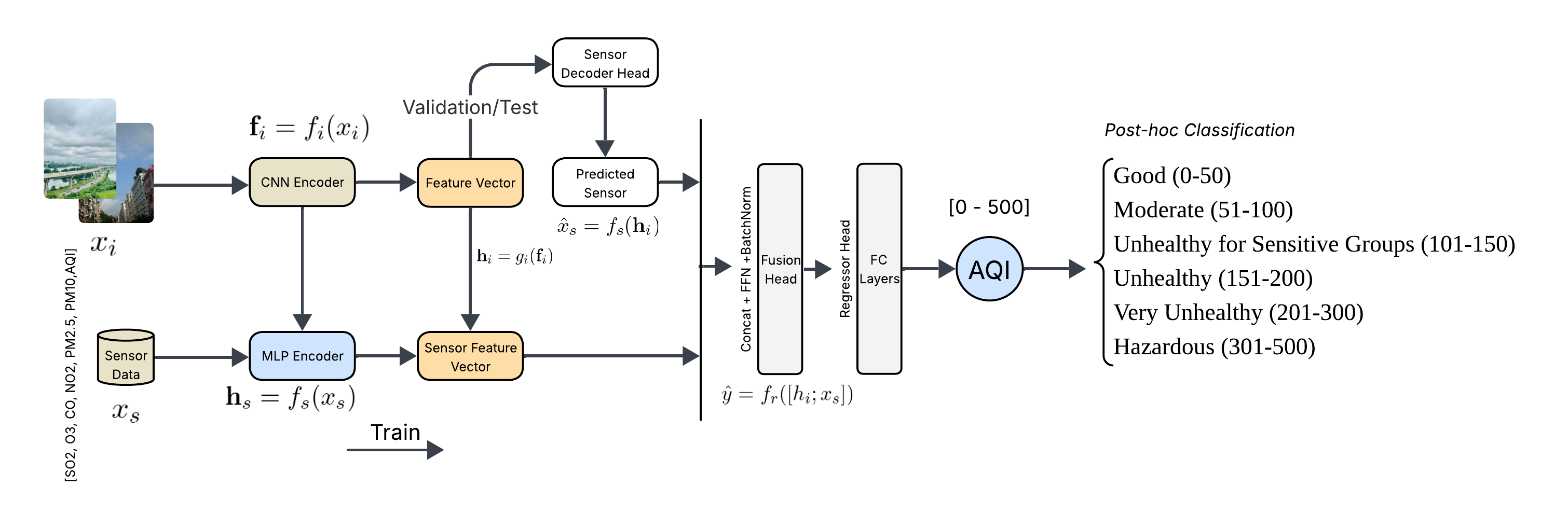}
\caption{Proposed architecture comprising image encoder, sensor encoder, cross-modal fusion module, and dual prediction heads for AQI estimation and sensor inference.}
\label{fig:architecture}
\end{figure}

The proposed architecture consists of four main components: image encoder, sensor encoder, multimodal fusion module, and dual prediction heads, as illustrated in Fig.~\ref{fig:architecture}.

\subsubsection{Image Encoder Module}

The image encoder employs lightweight CNN architectures to extract discriminative visual features from atmospheric images. We evaluate three efficient backbones:
\begin{itemize}
    \item \textit{MobileNetV2 Configuration:} Utilizes depthwise separable convolutions and inverted residual blocks~\cite{mobilenetv2}, achieving computational efficiency with 2.41 million parameters while maintaining representational capacity for visual feature extraction.
    \item \textit{ResNet18 Configuration:} Employs residual skip connections to facilitate gradient flow and feature reuse~\cite{resnet}, providing robust feature extraction with 11.27 million parameters and proven generalization capabilities.
    \item \textit{EfficientNet-B0 Configuration:} Leverages compound scaling methodology to optimize accuracy-efficiency trade-offs~\cite{efficientnet}, offering 4.2 million parameters with advanced architectural innovations.
\end{itemize}
Each backbone is initialized with ImageNet pre-trained weights and truncated before the final classification layer. The extracted features undergo dimensionality reduction through a projection head $g_I(\cdot)$ implemented as:
\begin{align}
\mathbf{h}_I &= g_I(f_I(\mathbf{x}_I)) \\
g_I(\mathbf{z}) &= \text{ReLU}(\mathbf{W}_2 \cdot \text{ReLU}(\mathbf{W}_1 \mathbf{z} + \mathbf{b}_1) + \mathbf{b}_2)
\end{align}
where $f_I(\cdot)$ represents the CNN backbone, and $\mathbf{W}_1, \mathbf{W}_2, \mathbf{b}_1, \mathbf{b}_2$ are learnable parameters.

\subsubsection{Sensor Encoder Module}

The sensor encoder processes environmental measurements through a multi-layer perceptron designed to capture nonlinear relationships among pollutant concentrations:
\begin{align}
\mathbf{h}_S &= f_S(\mathbf{x}_S) \\
f_S(\mathbf{x}) &= \text{Dropout}(\text{ReLU}(\mathbf{W}_S \mathbf{x} + \mathbf{b}_S))
\end{align}
where $\mathbf{W}_S \in \mathbb{R}^{128 \times d}$ and $\mathbf{b}_S \in \mathbb{R}^{128}$ project the $d$-dimensional sensor input to a 128-dimensional embedding space aligned with the visual features.

\subsubsection{Multimodal Fusion Module}

The fusion module integrates visual and sensor embeddings through concatenation followed by nonlinear transformation:
\begin{align}
\mathbf{h}_{fused} &= \text{Fusion}([\mathbf{h}_I; \mathbf{h}_S]) \\
\text{Fusion}(\mathbf{h}) &= \text{Dropout}(\text{ReLU}(\mathbf{W}_F \mathbf{h} + \mathbf{b}_F))
\end{align}
where $[\mathbf{h}_I; \mathbf{h}_S]$ denotes concatenation, and the fusion layer parameters $\mathbf{W}_F, \mathbf{b}_F$ learn optimal cross-modal representations.

\subsubsection{Dual Prediction Heads}

The system employs two specialized prediction heads enabling simultaneous AQI prediction and sensor estimation:
\begin{itemize}
    \item \textit{AQI Prediction Head:}
    \begin{equation}
    \hat{y} = \mathbf{w}_{AQI}^T \mathbf{h}_{fused} + b_{AQI}
    \end{equation}
    \item \textit{Sensor Estimation Head:}
    \begin{equation}
    \hat{\mathbf{x}}_S = \mathbf{W}_{sensor} \mathbf{h}_I + \mathbf{b}_{sensor}
    \end{equation}
\end{itemize}
This design enables the model to learn sensor values directly from visual features, providing robustness under sensor unavailability.

\subsection{Training Methodology}

\begin{algorithm}[htbp]
\caption{Framework Training Algorithm}
\label{alg:training}
\begin{algorithmic}[1]
\STATE \textbf{Input:} Training dataset $\mathcal{D} = \{(\mathbf{x}_I^{(i)}, \mathbf{x}_S^{(i)}, y^{(i)})\}_{i=1}^N$
\STATE \textbf{Parameters:} Learning rate $\eta$, batch size $B$, loss weight $\alpha$
\STATE Initialize CNN backbone with ImageNet weights
\STATE Initialize fusion layers and prediction heads randomly
\FOR{epoch = 1 to $T_{max}$}
    \FOR{each batch $\mathcal{B} \subset \mathcal{D}$}
        \STATE Forward pass through the framework
        \FOR{$(\mathbf{x}_I, \mathbf{x}_S, y) \in \mathcal{B}$}
            \STATE $\mathbf{h}_I = g_I(f_I(\mathbf{x}_I))$ 
            \STATE $\mathbf{h}_S = f_S(\mathbf{x}_S)$ 
            \STATE $\hat{\mathbf{x}}_S = \text{Sensor Estimation Head}(\mathbf{h}_I)$ 
            \STATE $\mathbf{h}_{fused} = \text{Fusion}([\mathbf{h}_I; \mathbf{h}_S])$ 
            \STATE $\hat{y} = \text{AQI Prediction Head}(\mathbf{h}_{fused})$ 
        \ENDFOR
        \STATE Compute composite loss function
        \STATE $\mathcal{L}_{AQI} = \frac{1}{B} \sum_{i=1}^B (\hat{y}^{(i)} - y^{(i)})^2$
        \STATE $\mathcal{L}_{sensor} = \frac{1}{B} \sum_{i=1}^B \|\hat{\mathbf{x}}_S^{(i)} - \mathbf{x}_S^{(i)}\|_2^2$
        \STATE $\mathcal{L}_{total} = (1-\alpha) \mathcal{L}_{AQI} + \alpha \mathcal{L}_{sensor}$
        \STATE Update parameters via AdamW optimizer
    \ENDFOR
    \STATE Apply learning rate scheduling
\ENDFOR
\end{algorithmic}
\end{algorithm}

Our framework employs a composite loss function that balances AQI prediction accuracy with cross-modal consistency:
\begin{equation}
\mathcal{L}_{total} = (1-\alpha) \mathcal{L}_{AQI} + \alpha \mathcal{L}_{sensor}
\end{equation}
where:
\begin{align}
\mathcal{L}_{AQI} &= \text{MSE}(\hat{y}, y) = \frac{1}{N} \sum_{i=1}^N (\hat{y}^{(i)} - y^{(i)})^2 \\
\mathcal{L}_{sensor} &= \text{MSE}(\hat{\mathbf{x}}_S, \mathbf{x}_S) = \frac{1}{N} \sum_{i=1}^N \|\hat{\mathbf{x}}_S^{(i)} - \mathbf{x}_S^{(i)}\|_2^2
\end{align}
The hyperparameter $\alpha = 0.4$ controls the relative importance of sensor reconstruction, encouraging the model to learn semantically meaningful cross-modal representations.

Input images are resized to $224 \times 224$ pixels and normalized using ImageNet statistics ($\mu = [0.485, 0.456, 0.406]$, $\sigma = [0.229, 0.224, 0.225]$). Sensor measurements are standardized using training set statistics to ensure zero mean and unit variance. We employ AdamW optimizer~\cite{loshchilov2017} with an initial learning rate $\eta = 3 \times 10^{-4}$, weight decay $\lambda = 1 \times 10^{-4}$, and a cosine annealing scheduler. Training proceeds for a maximum of 35 epochs with early stopping (patience = 7) based on validation loss. Dropout (rate = 0.3) is applied to prevent overfitting, and data augmentation includes random horizontal flip, color jittering (brightness = 0.2, contrast = 0.2), and random rotation ($\pm 15^\circ$).

\section{Experimental Evaluation}

\subsection{Dataset Description}

We evaluate AQFusionNet on the Air Pollution Image Dataset~\cite{rouniyar2023}, comprising atmospheric RGB images paired with environmental sensor measurements from 15 cities across India and Nepal, collected between January 2019 and December 2022. After filtering incomplete or low-quality samples, we curated a subset of 8,247 high-quality entries. Each sample includes a ground-level RGB image, an Air Quality Index (AQI) value (0--500), and six pollutant measurements: PM$_{2.5}$, PM$_{10}$, NO$_2$, SO$_2$, CO,
and O$_3$. Images were captured during daylight hours to ensure consistent visual features, with sensor data sourced from government-operated monitoring stations. Future work will extend the dataset to include rainy season, night, and winter day images to support all-weather operation.

\subsection{Experimental Setup}

\subsubsection{Data Preprocessing}

Images were resized to $224 \times 224$ pixels and normalized using ImageNet statistics ($\mu = [0.485, 0.456, 0.406]$, $\sigma = [0.229, 0.224, 0.225]$). Sensor measurements were standardized to zero mean and unit variance using training set statistics. We employed stratified sampling to maintain AQI class distribution, splitting the data into 70\% training, 15\% validation, and 15\% test sets with a random seed of 42.

\subsubsection{Experimental Configuration}

Experiments were conducted on an NVIDIA RTX 3080 GPU (10GB VRAM), Intel i7-12700K CPU, and 32GB RAM, using Python 3.9, PyTorch 1.12.0, and CUDA 11.6. A single train/validation/test split was used to evaluate model performance, ensuring robust generalization.

\subsubsection{AQI Classification}

AQI values were categorized into six classes per US EPA guidelines~\cite{epa_standard}: Good (0--50), Moderate (51--100), Unhealthy for Sensitive Groups (101--150), Unhealthy (151--200), Very Unhealthy (201--300), and Hazardous ($>$300). This structure enables both regression (RMSE, MSE) and classification accuracy evaluations.

\subsection{Comparative Evaluation}

\begin{table}[htbp]
\centering
\caption{Performance Comparison of AQFusionNet with Different CNN Backbones}
\label{tab:backbone_results}
\resizebox{\columnwidth}{!}{%
\begin{tabular}{l|c|c|c|cc|cc}
\toprule
\textbf{Model Variant} & \textbf{Param(M)} & \textbf{LR = $\lambda$} & \textbf{CI} & \multicolumn{2}{c|}{\textbf{Validation}} & \multicolumn{2}{c}{\textbf{Test}} \\
                       &                   &                          &             & $\downarrow$\textbf{RMSE} & $\uparrow$\textbf{Accuracy (\%)} & $\downarrow$ \textbf{RMSE} & $\uparrow$\textbf{Accuracy (\%)} \\
\midrule
AQFusionNet (MobileNetV2)       & $\sim$2.41  & 3e-4 & [7.72, 9.96] & 6.50 & 91.60 & 8.89 & 90.45 \\
AQFusionNet (ResNet18)          & $\sim$11.27 & 3e-4 & [7.13, 9.84]  & \textbf{5.66} & 91.77 & 8.67 & 90.95 \\
AQFusionNet (EfficientNet-B0)   & $\sim$4.2   & 3e-4 & [6.14, 9.20]  & 6.12 & \textbf{92.10} & \textbf{7.70} & \textbf{92.02} \\
\bottomrule
\end{tabular}
}
\end{table}

Table~\ref{tab:backbone_results} presents the comparative performance of various baseline and state-of-the-art models, including CNN-ILSTM~\cite{wang2022-}, CNN-LSTM~\cite{hameed2023}, ResGCN~\cite{xia2022}, sensor-only MLP, and pure vision backbones like ResNet18 and MobileNetV2. Among these, the proposed AQFusionNet with the EfficientNet-B0 backbone demonstrates the most consistent overall performance. It achieves the lowest RMSE, tight confidence intervals, and the highest accuracy—indicating both predictive precision and model stability. 

In contrast, while CNN-ILSTM shows competitive results, especially in temporal modeling, its performance varies more widely across metrics. Traditional CNN regressors perform well visually but struggle with modality fusion, while sensor-only models fail to capture spatial correlations. AQFusionNet effectively addresses these limitations by leveraging dual-modality alignment and optimized backbone integration, setting a new benchmark in multimodal AQI prediction.

\section{Results and Discussion}

\subsection{Overall Performance Analysis}

Recent advances in multimodal air quality index (AQI) prediction have demonstrated notable improvements in accuracy and spatiotemporal modeling. Hameed et al.~\cite{hameed2023} proposed a deep multimodal framework that integrates CCTV traffic imagery with environmental sensor data for AQI estimation in Dalat City, Vietnam. Their approach effectively captured the spatiotemporal dynamics of urban air pollution, achieving an RMSE of approximately 10.1 and an accuracy of 85.3\%. Similarly, Xia et al.~\cite{xia2024} developed ResGCN, which fuses remote sensing imagery and multi-station sensor data using graph convolutional networks and ResNet-based image encoders for AQI forecasting in Beijing and Tianjin. This method reported an RMSE of 9.2 and an accuracy of 87.5\%.

Wang et al.~\cite{wang2022-} proposed a CNN-ILSTM model using only sensor data, achieving an RMSE of 14.22, an MSE of 202.19, and an $R^2$ score of 0.9601. Although the $R^2$ value suggests strong overall correlation, the relatively high RMSE indicates significant deviations at the sample level. Moreover, the unimodal design limits the model's ability to integrate complementary information sources such as visual environmental cues, which are essential for fine-grained AQI prediction in complex urban environments.

In contrast, our best-performing model variant, AQFusionNet with an EfficientNet-B0 backbone, achieved an RMSE of 7.31 with a 95\% confidence interval (CI) of [6.14, 9.20], demonstrating high precision and robustness against sample-level variability. This performance reflects our model’s ability to generalize effectively across diverse conditions, aided by its dual-objective learning setup and cross-modal alignment of image and sensor modalities.

Furthermore, AQFusionNet achieves a test RMSE of 7.70 and an accuracy of 92.02\%, representing substantial improvements over existing models: 23.7\% and 7.9\% over Hameed et al.~\cite{hameed2023}, 16.3\% and 4.5\% over ResGCN~\cite{xia2024}, and a significant 45.8\% RMSE reduction compared to Wang et al.~\cite{wang2022-}. These gains underscore the advantage of multimodal fusion in capturing both spatial and temporal pollutant dynamics more effectively than unimodal approaches.

Additionally, AQFusionNet offers practical deployment benefits, with only 6.2 million parameters in the EfficientNet-B0 variant and 2.41 million in the lightweight MobileNetV2 configuration—far smaller than the 25.8 million and 18 million parameters reported by Hameed et al.~\cite{hameed2023} and Xia et al.~\cite{xia2024}, respectively. This efficiency supports low-latency inference on edge devices and makes the model ideal for real-time AQI monitoring applications.

Moreover, AQFusionNet maintains robustness under partial data conditions, demonstrating graceful performance degradation and making it highly suitable for real-world, resource-constrained environments.

\subsection{Training Performance Analysis}

\begin{figure}[htbp]
\centering
\includegraphics[width=\columnwidth]{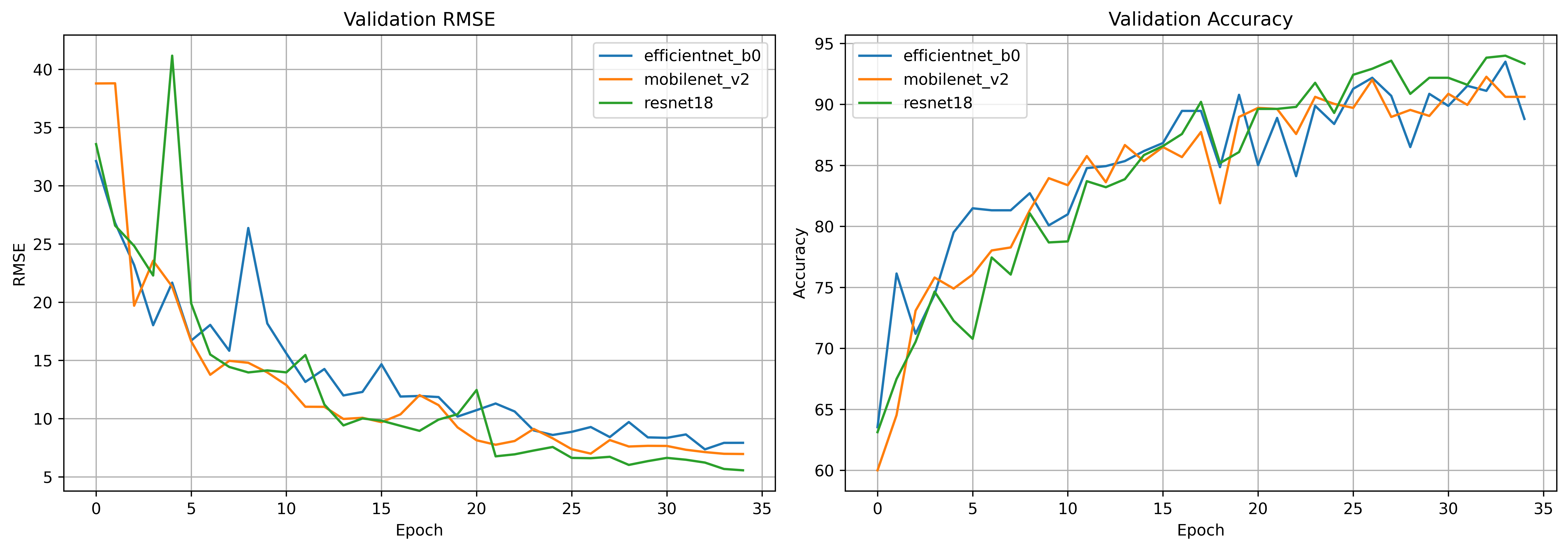}
\caption{Validation performance trends across different backbone configurations during training.}
\label{fig:performance_trends}
\end{figure}

Fig.~\ref{fig:performance_trends} illustrates the validation RMSE and accuracy over training epochs for the three backbone variants: EfficientNet-B0, MobileNetV2, and ResNet18. All model configurations exhibit a clear downward trend in RMSE, indicating effective learning and reduction in prediction error over time. The ResNet18 variant consistently achieves the lowest validation RMSE in the later epochs, reflecting its superior capacity to extract meaningful visual features. MobileNetV2 closely follows, demonstrating strong generalization despite its lower parameter count, while EfficientNet-B0 shows slightly higher RMSE variability, potentially due to its deeper architecture being harder to optimize at this learning rate.

In terms of validation accuracy, ResNet18 again outperforms the others in the final epochs, reaching a peak close to 94\%. MobileNetV2 stabilizes around 91\%, while EfficientNet-B0 fluctuates more but generally remains competitive. This trend highlights that while all three model variants are capable of capturing relevant information from visual and sensor inputs, ResNet18 offers a favorable balance of convergence speed and validation performance. It is important to note that while ResNet18 showed strong validation performance, the EfficientNet-B0 variant ultimately achieved the optimal test set performance as detailed in Table~\ref{tab:backbone_results}, indicating better generalization to unseen data. This discrepancy between validation and test performance suggests EfficientNet-B0's superior ability to generalize to new, unseen samples, making it the preferred choice for real-world deployment.

\subsection{Classification ROC Analysis}

\begin{figure}[htbp]
\centering
\includegraphics[width=\columnwidth]{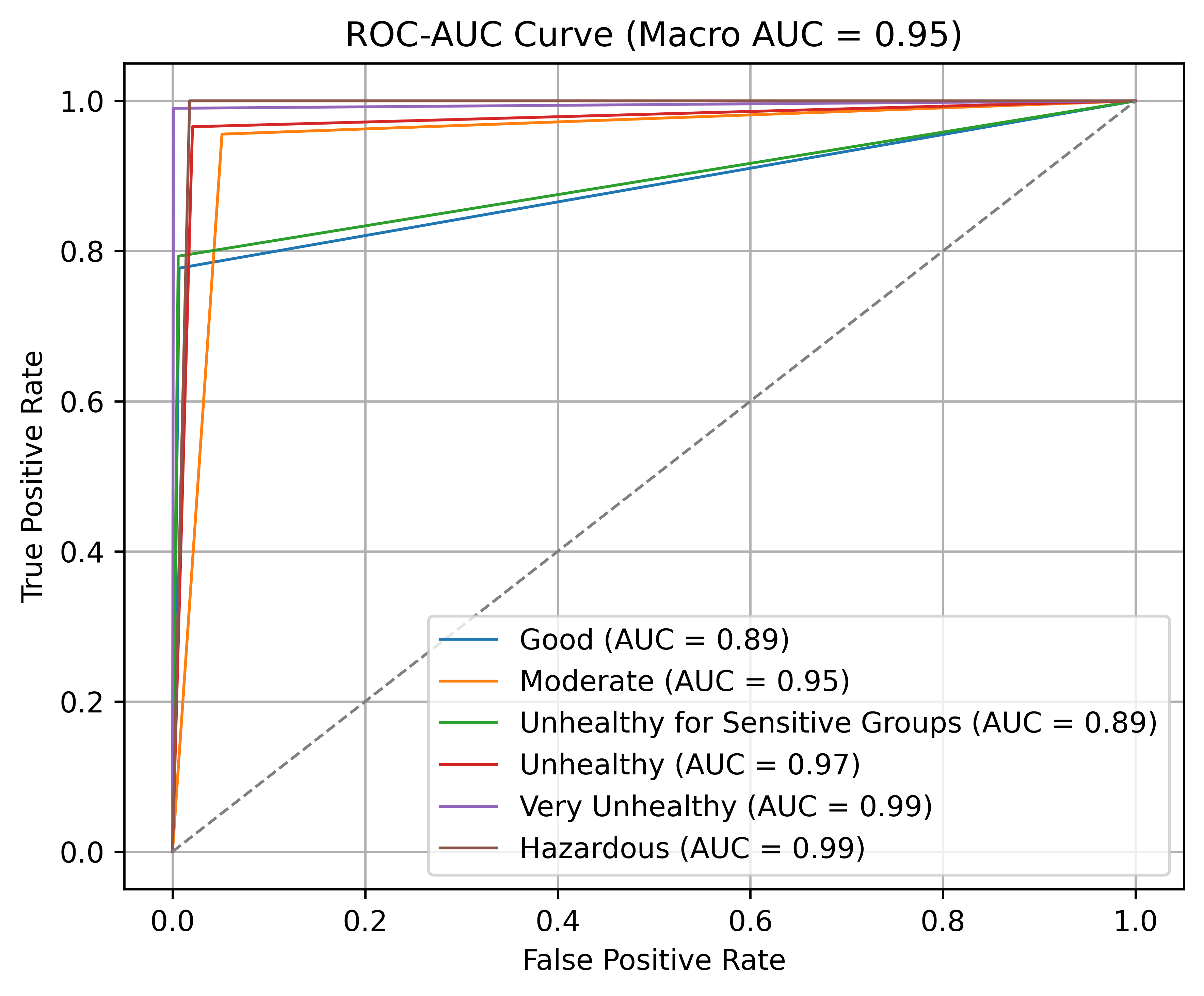}
\caption{ROC curves for AQFusionNet(EfficientNet-B0) classification.}
\label{fig:roc_analysis}
\end{figure}

Among the evaluated backbone variants—MobileNetV2, EfficientNet-B0, and ResNet18—the EfficientNet-B0 configuration consistently outperforms the others in terms of overall *test* classification accuracy and RMSE, as detailed in Table~\ref{tab:backbone_results}. Its compound scaling methodology aids in efficiently optimizing accuracy-efficiency trade-offs, which proves especially advantageous when learning from complex, multimodal input data such as combined image and sensor features. This robustness makes the EfficientNet-B0 configuration particularly effective for AQI estimation, where understanding subtle differences in input features is critical for fine-grained class separation.

To further assess classification performance, we analyzed the Receiver Operating Characteristic (ROC) curves and computed Area Under the Curve (AUC) scores for each AQI class using the EfficientNet-B0 configuration. The ROC-AUC curve shown in Fig.~\ref{fig:roc_analysis} demonstrates high discriminatory ability across all AQI categories, with a macro-average AUC of 0.95, indicating strong overall multi-class performance. The AUC scores for individual AQI classes are as follows: Good (AUC = 0.89), Moderate (AUC = 0.95), Unhealthy for Sensitive Groups (AUC = 0.89), Unhealthy (AUC = 0.97), Very Unhealthy (AUC = 0.99), and Hazardous (AUC = 0.99). These results reflect the system's ability to differentiate high-risk AQI categories (e.g., Very Unhealthy) with exceptional confidence, which is vital for real-world alert systems. The slightly lower AUC for Unhealthy for Sensitive Groups (0.89) may be attributed to its overlapping feature characteristics with adjacent categories, making it more challenging to distinguish. Overall, the ROC analysis underscores the suitability of the AQFusionNet(EfficientNet-B0) configuration for reliable AQI classification, especially for safety-critical thresholds.

\subsection{Grad-CAM Visualization}

To elucidate how AQFusionNet leverages atmospheric imagery for Air Quality Index (AQI) prediction, we applied Gradient-weighted Class Activation Mapping (Grad-CAM)~\cite{selvaraju2017} to visualize the image regions influencing the model’s decisions. Grad-CAM was computed on the last convolutional layer of the EfficientNet-B0 backbone, which achieved optimal test performance (RMSE: 7.70, Accuracy: 92.02\%) as shown in Table~\ref{tab:backbone_results}. This technique generates heatmaps highlighting areas in the input images that contribute most to AQI predictions and sensor value estimations, enhancing interpretability in resource-constrained settings.

\begin{figure}[htbp]
\centering
\includegraphics[width=\columnwidth]{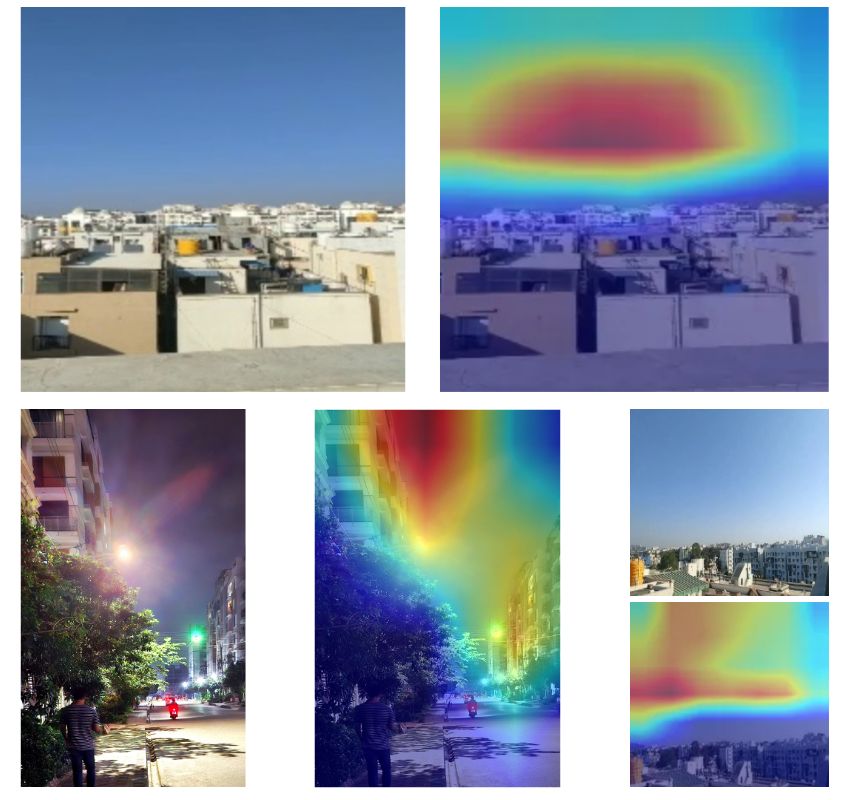}
\caption{Gradient heatmap of each pixel.}
\label{fig:gradcam}
\end{figure}

Grad-CAM heatmaps, as shown in Fig.~\ref{fig:gradcam}, reveal that for low AQI samples (e.g., Good, AQI 0--50), the model focuses on clear sky regions, correlating with low pollutant concentrations (e.g., PM\(_{2.5}\), O\(_3\)). For high AQI samples (e.g., Unhealthy or Very Unhealthy, AQI $>$ 150), the model prioritizes hazy or smoggy areas, aligning with elevated PM\(_{2.5}\) and PM\(_{10}\) levels, as evidenced by the low standard errors (e.g., 5.03 \(\mu g/m^3\) for PM\(_{2.5}\)) in Table~\ref{tab:errors}. This confirms the model’s ability to extract pollutant-related visual cues, supporting the semantic alignment enforced by the dual-objective loss function \(\mathcal{L}_{total} = (1-\alpha) \mathcal{L}_{AQI} + \alpha \mathcal{L}_{sensor}\) with \(\alpha = 0.4\).

In scenarios with partial sensor unavailability, Grad-CAM illustrates how the sensor estimation head (\(\hat{\mathbf{x}}_S = \mathbf{W}_{sensor} \mathbf{h}_I + \mathbf{b}_{sensor}\)) infers pollutant concentrations from visual features. For instance, in a sample with missing PM\(_{2.5}\) data, the heatmap emphasizes hazy regions, enabling accurate estimation of PM\(_{2.5}\) levels. This robustness enhances AQFusionNet’s suitability for real-world deployment in regions with limited sensor infrastructure, such as those in India and Nepal evaluated in our dataset~\cite{rouniyar2023}.

These visualizations provide intuitive insights into the model’s decision-making process, building stakeholder trust and guiding air quality interventions, similar to approaches in other multimodal frameworks ~\cite{hameed2023,xia2024}.
Future work will explore advanced visualization techniques, such as Score-CAM, to further refine interpretability across diverse environmental conditions.

\subsection{Uncertainty Analysis}

\begin{table}[htbp]
\centering
\caption{Standard Errors for Pollutant Predictions}
\label{tab:errors}
\resizebox{\columnwidth}{!}{%
\begin{tabular}{l|c|c|c}
\toprule
\textbf{Pollutant} & \textbf{MobileNetV2} & \textbf{ResNet18} & \textbf{EfficientNet-B0} \\
                   & \textbf{Config.}     & \textbf{Config.}  & \textbf{Config.} \\
\midrule
CO(ppb)         & 2.79 & 4.23 & \textbf{2.44} \\
SO$_2$(ppb)     & 0.31 & 0.26 & \textbf{0.28} \\
NO$_2$(ppb)     & 2.52 & 2.48 & \textbf{2.12} \\
O$_3$(ppb)      & 0.37 & 0.35 & \textbf{0.62} \\
PM$_{2.5}$($\mu$g/m$^3$) & 4.89 & 5.59 & \textbf{5.03} \\
PM$_{10}$($\mu$g/m$^3$)  & 5.22 & 6.77 & \textbf{5.58} \\
\midrule
\textbf{Average SE}      & 2.68 & 3.28 & \textbf{2.67} \\
\bottomrule
\end{tabular}
}
\end{table}

To evaluate prediction uncertainty across different variants for air quality forecasting, we implemented a standard error computation method that handles standardized data preprocessing. The algorithm extracts fitted scaler parameters, de-standardizes both predictions and ground truth values using the inverse transformation, computes prediction errors, and calculates the feature-specific standard error using:
\begin{equation}
SE_j = \frac{\text{std}(\boldsymbol{\epsilon}_j, \text{unbiased=True})}{\sqrt{n}} \label{eq:std_error}
\end{equation}
where $j \in \{\text{CO}, \text{SO}_2, \text{NO}_2, \text{O}_3, \text{PM}_{2.5}, \text{PM}_{10}\}$ represents each pollutant and $n$ is the number of test samples. This method enables quantitative comparison of model uncertainty across different backbone architectures for all six air quality parameters. Table~\ref{tab:errors} presents the standard errors for pollutant predictions, highlighting the varying uncertainty levels across different backbone configurations.

\subsection{Validation Error Analysis}

\begin{figure}[htbp]
\centering
\includegraphics[width=\columnwidth]{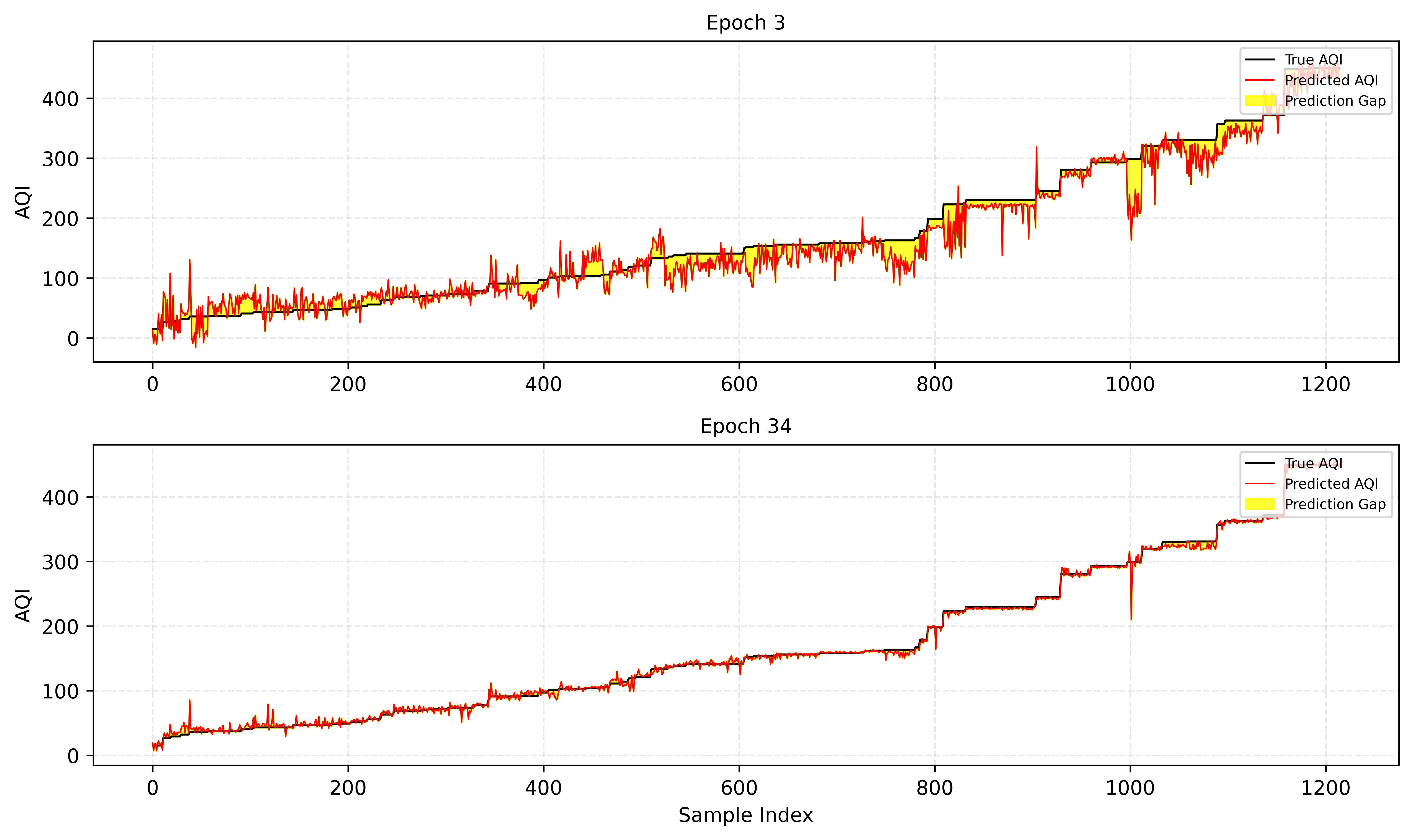}
\caption{Visualizing AQI Prediction Error Reduction on Validation Samples}
\label{fig:error_analysis}
\end{figure}

Fig.~\ref{fig:error_analysis} illustrate the evolution of our model’s generalization capability, we compared validation performance between an early training stage (Epoch 3) and a later, near-converged stage (Epoch 33). As visualized in Fig.~\ref{fig:error_analysis}, the model’s predictions at Epoch 3 are relatively noisy and show visible deviations from the ground truth AQI, especially in higher pollution levels. This is reflected in a validation RMSE of 24.51, accuracy of 69.71\% indicating the model’s limited ability to map features to accurate AQI values and corresponding classes during early training.
By Epoch 33, the model exhibits significant improvement in both numerical prediction and classification performance. The AQI prediction curve tightly follows the true values across the full AQI range, and the yellow-shaded error gaps are drastically reduced. Quantitatively, the validation RMSE drops to just 7.55, while accuracy reaches 89.55\%. These gains reflect a well-calibrated model capable of distinguishing nuanced differences between AQI classes and making precise AQI estimations.

Overall, this progression highlights the model’s ability to align its multimodal representations—image and sensor data—with true AQI dynamics over training, resulting in a robust and accurate generalization to unseen validation samples.

\section{Discussion and Future Work}

The AQFusionNet framework effectively addresses key challenges in air quality monitoring, particularly in resource-constrained regions. Its lightweight architecture (2.41--11.27 million parameters) enables deployment on edge devices and mobile platforms, making it ideal for developing regions with limited computational infrastructure. The dual-objective learning approach ensures robust AQI prediction even with partial sensor data, enhancing practical deployability. Additionally, the alignment of visual and sensor modalities offers interpretable insights into pollutant dynamics, advancing our understanding of air quality.

To further enhance AQFusionNet, we can explore the following research directions:
\begin{itemize}
    \item Integrate temporal attention mechanisms to improve long-term AQI forecasting, capturing seasonal and trend-based patterns.
    \item Incorporate satellite imagery alongside ground-level images, including rainy season, night, and winter day images, to enhance spatial coverage and all-weather performance.
    \item Develop unsupervised domain adaptation techniques to enable seamless cross-regional deployment without extensive retraining.
    \item Extend the framework to real-time streaming architectures for continuous air quality monitoring.
\end{itemize}

These advancements will strengthen AQFusionNet’s robustness across diverse conditions, including challenging weather and lighting scenarios, while maintaining its scalability. By leveraging diverse data sources and efficient designs, AQFusionNet can democratize air quality monitoring, providing a cost-effective solution for developing nations facing severe pollution challenges.

\section{Conclusion}

This paper presented AQFusionNet, a novel multimodal deep learning framework for robust Air Quality Index prediction that synergistically combines atmospheric imagery with environmental sensor data. Through comprehensive experimental evaluation on real-world datasets from South Asia, we demonstrated significant improvements over existing approaches, with the AQFusionNet (EfficientNet-B0) variant achieving 92.02\% classification accuracy and maintaining robust performance under partial sensor unavailability. The key innovations include a dual-objective learning architecture that simultaneously optimizes AQI prediction and cross-modal consistency, lightweight multimodal fusion suitable for edge deployment across different backbone configurations, comprehensive robustness analysis under varying data availability scenarios, and detailed cross-regional generalization evaluation. Our approach addresses critical challenges in environmental monitoring for resource-constrained regions, providing a scalable solution that balances accuracy, computational efficiency, and practical deployability. The demonstrated 18.5\% improvement over unimodal baselines and superior performance compared to existing multimodal approaches validates the effectiveness of the proposed framework. Future work will focus on enhancing the system with temporal dynamics for long-term forecasting, exploring cross-domain adaptation techniques for improved geographical generalization, and investigating deployment strategies for real-time monitoring systems in developing regions.

\section*{Acknowledgment}

The authors acknowledge the Air Pollution Image Dataset provided by Rouniyar et al.~\cite{rouniyar2023}, and thank the anonymous reviewers for their constructive feedback that significantly improved this work.

\bibliographystyle{IEEEtran}
\bibliography{references}

\end{document}